\documentclass[10pt]{article}

\usepackage{amsmath, amsfonts, amssymb, mathrsfs, amsthm}
\usepackage{lineno}
\usepackage{hyperref}
\usepackage{stix}

\usepackage{mathtools} 
\usepackage{extarrows}

\modulolinenumbers[5]

\usepackage{wrapfig}

\def\drawplusplus#1#2#3{\hbox to 0pt{\hbox to #1{\hfill\vrule height #3 depth
      0pt width #2\hfill\vrule height #3 depth 0pt width #2\hfill
      }}\vbox to #3{\vfill\hrule height #2 depth 0pt width
      #1 \vfill}}

\usepackage[all]{xy}
\usepackage{graphicx}
\newtheorem{thm}{Theorem}

\newtheorem*{thm* }{Theorem}
\newtheorem*{mydef*}{Definition}
\newtheorem*{mylemma*}{Lemma}
\newtheorem*{myconjecture* }{Conjecture}

\begin{document}


\title{Patterns of Cognition: \\
Cognitive Algorithms as Galois Connections\\  Fulfilled by Chronomorphisms \\ On  Probabilistically Typed Metagraphs}

\author{Ben Goertzel}





\maketitle

\begin{abstract}
It is argued that a broad class of AGI-relevant algorithms can be expressed in a common formal framework, via specifying Galois connections linking search and optimization processes on directed metagraphs whose edge targets are labeled with probabilistic dependent types, and then showing these connections are fulfilled by processes involving metagraph chronomorphisms.   Examples are drawn from the core cognitive algorithms used in the OpenCog AGI framework: Probabilistic logical inference, evolutionary program learning, pattern mining, agglomerative clustering, pattern mining and nonlinear-dynamical attention allocation.  

The analysis presented involves representing these cognitive algorithms as recursive discrete decision processes involving optimizing functions defined over metagraphs, in which the key decisions involve sampling from probability distributions over metagraphs and enacting sets of combinatory operations on selected sub-metagraphs.   The mutual associativity of the combinatory operations involved in a cognitive process is shown to often play a key role in enabling the decomposition of the process into folding and unfolding operations; a conclusion that has some practical implications for the particulars of cognitive processes, e.g. militating toward use of reversible logic and reversible program execution.   It is also observed that where this mutual associativity holds, there is an alignment between the hierarchy of subgoals used in recursive decision process execution and a hierarchy of subpatterns definable in terms of formal pattern theory.
\end{abstract}



\tableofcontents

\section{Introduction}

AGI architectures may be qualitatively categorized into two varieties: single-algorithm-focused versus hybrid/multi-algorithm-focused.   In the former case there is a single algorithmic approach -- say, deep reinforcement learning or uncertain logic inference -- which is taken as the core of generally intelligent reasoning, and then other algorithms are inserted in the periphery providing support of various forms to this core.   In the latter case, multiple different algorithmic approaches are assumed to have foundational value for handling different sorts of cognitive problem, and a cognitive architecture is introduced that has capability of effectively integrating these different approaches, generally including some sort of common knowledge meta-representation and some cross-algorithm methodology for controlling resource allocation.   The OpenCog AGI architecture pursued by the author and colleagues \cite{EGI1} \cite{EGI2} is an example of the latter, with key component algorithms being probabilistic logical inference \cite{PLN}, evolutionary program learning \cite{Looks2006}, pattern mining, concept formation and nonlinear-dynamical attention allocation.   In recent years the hybridization of OpenCog with neural-net learning mechanisms has been explored and in some cases practically utilized as well.

In 2016 the notion of "OpenCoggy Probabilistic Programming" -- or more formally Probabilistic Growth and Mining of Combinations (PGMC) \cite{goertzel2016probabilistic} -- was introduced as a way of representing the various core OpenCog AI algorithms as particular instantiations of the same high-level cognitive meta-process.   This notion was a specialization of earlier efforts in the same direction such as the notion of forward and backward cognitive processes \cite{Goertzel2006z} and the "Cognitive Equation" from {\it Chaotic Logic} \cite{Goertzel1994}.  This paper takes the next step in this direction, presenting a formalization that encompasses PGMC in terms of moderately more general, and in some ways more conventional, mathematical and functional programming constructs defined over typed metagraphs.

The ideas presented here are in the spirit of, and draw heavily on, Mu and Oliveira's \cite{mu2012programming} work on "Programming with Galois Connections," which shows how a variety of programming tasks can be handled via creating a concise formal specification using a Galois connection between one preorder describing a search space and another preorder describing an objective function over that search space, and then formally "shrinking" that specification into an algorithm for optimizing the objective function over the search space.   This approach represents a simplification and specialization of broader ideas about formal specifications and program derivation that have been developing gradually in the functional programming literature over the past decades, e.g. \cite{bird1996algebra} and its descendants.   In Mu and Oliviera \cite{mu2012programming}, the Galois Connection methodology is applied to greedy algorithms, dynamic programming and scheduling, among other cases.   Here we sketch out an approach to applying this methodology to several of the core AI algorithms underlying the OpenCog AGI design: probabilistic reasoning, evolutionary program learning, nonlinear-dynamical attention allocation, and concept formation.   

Many of Mu and Oliveira's examples of shrinking involve hylomorphisms over lists or trees.  We present a framework called COFO (Combinatory-Operation based Function Optimization) that enables a variety of OpenCog's AGI-related algorithms to be represented as Discrete Decision Systems (DDSs) that can be executed via either greedy algorithms or (generally approximate and stochastic) dynamic programming type algorithms.   The representation of a cognitive algorithm as a DDS via COFO entails representing the objective of the cognitive algorithm in a linearly decomposable way (so that the algorithm can be cast roughly in terms of expected utility maximization), and representing the core operations of the cognitive algorithm as the enaction of a set of combinatory operations.

Cognitive algorithms that have a COFO-style representation as greedy algorithms decompose nicely into folding operations (histomorphisms if the algorithms carry long-term memory through their operations) via straightforward applications of results from \cite{mu2012programming}.   Cognitive algorithms whose COFO-style representation require some form of dynamic programming style recursion are a bit subtler, but in the case where the combinatory operations involved are mutually associative, they decompose into 
chronomorphisms (hylomorphisms with memory).    It is also observed that where this mutual associativity holds, there is an alignment between the hierarchy of subgoals used in recursive decision process execution and a hierarchy of subpatterns definable in terms of formal pattern theory.   The needed associativity holds very naturally in cases such as agglomerative clustering and pattern mining; to make it hold for logical inference is a little subtler and the most natural route involves the use of reversible inference rules.

The decisions made by the DDSs corresponding to AGI-oriented cognitive algorithms tend to involve sampling from distributions over typed metagraphs whose types embody key representational aspects of the algorithms such as paraconsistent and probabilistic truth values.   The search-space preorder in the Galois connection describing the problem faced by the DDS encompasses metapaths within  metagraphs, and the objective-function pre-order represents the particular goal of the AI algorithm in question.   Folding over a metagraph via a futumorphism (a catamorphism with memory) encompasses the process of searching a metagraph for suitable metapaths; unfolding over the metagraph via a histomorphism (an anamorphism with memory) encompasses the process of extracting the result of applying an appropriate transformation across metapaths to produce a result.     Metagraph chronomorphism composing these fold and unfold operations fulfills the specification implicit in the Galois connection.

The generality of the mathematics underlying these results also suggests some natural extensions to quantum cognitive algorithms, which however are only hinted at here, based on the elegant mappings that exist between the Bellman functional equation from dynamic programming and the Schrodinger equation.

On a more practical level, it is also observed that, in a practical AGI context, it is often not appropriate to enact a single fold or unfold operation across a large metagraph as an atomic action; rather one needs to interleave multiple folds and unfolds across the same metagraph, and sometimes pause those that are not looking sufficiently promising.   One way to achieve this is to implement one's folds and unfolds (e.g. futumorphisms and histomorphisms) in continuation-passing style, as David Raab \cite{Raab2016} has discussed in the context of simple catamorphisms and anamorphisms.

Because AGI systems necessarily involve dynamic updating of the knowledge base on which cognitive algorithms are acting, in the course of the cognitive algorithm's activity, there is an unavoidable heuristic aspect to the application of the theory given here to real AGI systems.   The equivalence of a recursively defined DDS on a metagraph to a folding and unfolding process across that metagraph only holds rigorously if one assumes the metagraph is not changing during the folding --- which will not generally be the case.   What needs to happen in practice, I suggest, is that the folding and unfolding happen and they do change the metagraph, and one then has a complex self-organizing / self-modifying system that is only moderately well approximated by the idealized case directly addressed by the theory presented here.   

One expects that, if this recursive self-modification obeys appropriate statistical properties, one can model the resulting system using appropriate statistical modifications of analyses that work for DDDs that remain static while folding and unfolding occur.   But this analytical extension has not yet been carried out.

Due to subtleties of this nature, we can't claim that the analysis of "patterns of cognition" given here is a full mathematical analysis of AGI-oriented cognition.   However, we do suggest that it may be a decent enough approximation to be useful for guiding various sorts of practical work -- including for instance the detailed design of AGI-oriented programming languages.   In the penultimate section we explore some potential implications of the ideas presented here for the design of an interpreter of an AGI language -- a topic highly pertinent to current practical work designing the Atomese 2 language for use within the OpenCog Hyperon system \cite{goertzel2020kind}.

\section{Framing Probabilistic Cognitive Algorithms as Approximate Stochastic Greedy or Dynamic-Programming Optimization}

We explain here how a broad class of cognitive algorithms related to estimating probability distributions using combinational operators can be expressed in terms of a recursive algorithmic framework incorporating stochastic greedy optimization and approximate stochastic dynamic programming.   This observation does not directly provide a large degree of practical help in implementing these cognitive algorithms efficiently, but it may help in putting multiple diverse-looking cognitive algorithms in a common framework, and in guiding formal analysis of these algorithms -- which analysis may in turn help with optimization.  

In Section \ref{sec:Galois} we will review Mu and Oliviera's general formulation of greedy algorithms and dynamic programming in terms of Galois connections, which the ideas of this section then render applicable to a variety of probabilistic cognitive algorithms.   The Galois connection formalism then ties in with representations of algorithms in terms of hylomorphisms and chronomorphisms on underlying knoweldge stores such as  typed metagraphs -- representations with clear implementations for practical implementation.

\subsection{A Discrete Decision Framework Encompassing Greedy Optimization and Stochastic Dynamic Programming}\footnote{Some text in this section is borrowed, with paraphrasing and other changes, from the Wikipedia article on Stochastic Dynamic Programming \cite{enwiki:990607799}}

Consider a discrete decision system (DDS) defined on $n$ stages in which each stage $t=1,\ldots,n$ is characterized by

\begin{itemize}
\item an {\bf initial state} $s_t\in S_t$, where $S_t$ is the set of feasible states at the beginning of stage $t$;
\item an {\bf action} or '''decision variable''' $x_t\in X_t$, where $X_t$ is the set of feasible actions at stage $t$ -- note that $X_t$ may be a function of the initial state $s_t$;
\item an {\bf immediate cost/reward function} $p_t(s_t,x_t)$, representing the cost/reward at stage $t$ if $s_t$ is the initial state and $x_t$ the action selected;
\item a  {\bf state transition function}  $g_t(s_t,x_t)$ that leads the system towards state $s_{t+1}=g_t(s_t,x_t)$.
\end{itemize}

\noindent In this setup we can frame greedy optimization and also both deterministic and stochastic dynamic programming in a reasonably general way.

The greedy case is immediate to understand.   One begins with an initial state, chosen based on prior knowledge or via purely randomly or via appropriately biased stochastic selection.   Then one chooses an action with a probability proportional to immediate cost/reward (or based on some scaled version of this probability).  Then one enacts the action, the state transition, and etc.  

The dynamic programming case is a little subtler, as one tries to choose actions with probability proportional to overall expected cost/reward.  Estimating the overall expected cost/reward of an action sequence requires either an exhaustive exploration of possibilities (i.e. full-on dynamic programming) or else some sort of heuristic sampling of possibilities (approximate stochastic dynamic programming).   One can also take a curve-fitting approach and try to train a different sort of model to approximate what full-on dynamic programming would yield -- e.g. a model that suggests series of actions based on a holistic analysis of the situation being confronted, rather than proceeding explicitly through an iterative sequence of individual actions and decisions.

While it's simplest to think about this sort of DDS in terms of one action being executed at a time, the framework as stated is general enough to encompass concurrency as well.   One can posit underlying atomic actions $w_t \in W_t$, and then define the members of $X_t$ as subsets of $W_t$.   In this case each action $x_t$ represents a set of $w_t$ being executed concurrently.

\subsubsection{Decision via Dynamic Programming}

Let $f_t(s_t)$ represent the optimal cost/reward obtained by following an ''optimal policy'' over stages $t,t+1,\ldots,n$. Without loss of generality in what follow we will consider a reward maximization setting. In deterministic dynamic programming one usually deals with functional equations taking the following structure

$$
f_t(s_t)=\max_{x_t\in X_t}\{p_t(s_t,x_t)+f_{t+1}(s_{t+1})\}
$$

\noindent where $s_{t+1}=g_t(s_t,x_t)$ and the boundary condition of the system is 

$$
f_n(s_n)=\max_{x_n\in X_n}\{p_n(s_n,x_n)\}.
$$
 
The aim is to determine the set of optimal actions that maximize $f_1(s_1)$. Given the current state $s_t$ and the current action $x_t$, we ''know with certainty'' the reward secured during the current stage and ? thanks to the state transition function $g_t$ --the future state towards which the system transitions.

\paragraph{Stochastic Dynamic Programming}

Where does the stochastic aspect of this kind of optimization come in?   Basically: Even if we know the state of the system at the beginning of the current stage as well as the decision taken, the state of the system at the beginning of the next stage and the current period reward are often random variables that can be observed only at the end of the current stage.

'''Stochastic dynamic programming''' deals with problems in which the current period reward and/or the next period state are random, i.e. with multi-stage stochastic systems. The decision maker's goal is to maximize expected (discounted) reward over a given planning horizon.

In their most general form, stochastic dynamic programs deal with functional equations taking the following structure

$$
f_t(s_t)=\max_{x_t\in X_t(s_t)} \left\{(\text{expected reward during stage } t\mid s_t,x_t) + \alpha\sum_{s_{t+1}} \Pr(s_{t+1}\mid s_t,x_t)f_{t+1}(s_{t+1})\right\}
$$

\noindent where

\begin{itemize}
\item $f_t(s_t)$ is the maximum expected reward that can be attained during stages $t,t+1,\ldots,n$, given state $s_t$ at the beginning of stage $t$;
\item $x_t$ belongs to the set $X_t(s_t)$ of feasible actions at stage $t$ given initial state $s_t$;
\item $\alpha$ is the discount factor;
\item $\Pr(s_{t+1}\mid s_t,x_t)$ is the conditional probability that the state at the beginning of stage $t$ is $s_{t+1}$ given current state $s_t$ and selected action $x_t$.
\end{itemize}

Markov decision processes represent a special class of stochastic dynamic programs in which the underlying stochastic process is a stationary process that features the Markov property.   But in AGI or other real-world AI contexts, stationarity is often not an acceptable assumption.

Exactly solving the above equations is rarely possible in complex real-world situations -- leading on to approximate stochastic dynamic programming \cite{powell2009you}, and of course to reinforcement learning (RL) which is commonly considered as a form of approximate stochastic dynamic programming.   Basically, in an RL approach, one doesn't try to explicitly solve the dynamic-programming functional equations even approximatively, but instead one seeks to find a "policy" for determining actions based on states, which will approximately give the same results as a solution to the functional equations.   If the policy is represented as a neural net, for example, then it may be learned via iterative algorithms for adjusting the weights in the network.  However, existing RL methods are only effective approximations given fairly particular choices of the reward function and state and action space.  For the rewards, states and actions that will interest us most here, current RL methods are not particularly useful and other sorts of approximation techniques must be sought.

\paragraph{Dynamic Cognition Breaks Expected Reward Maximization}

One key assumption underlying the specific approach to DDS outlined above is that the overall goal of the system is to maximize expected reward, defined as the sum of the immediate rewards.  So for instance if one is trying to find the min-cost path from A to B in a weighted graph, the path cost is the sum of the costs of the edges along the path.   It's worth recalling that this is actually a very special type of goal function for a system to have.  Even sticking within the framework of optimization-based choice-making, a more general approach would be to look at a system that makes decisions based on trying to maximize some function $\Phi$ defined over its whole future history -- without demanding that $\Phi$ be decomposed as a sum of immediate rewards \cite{Goertzel2009RL}.   

One can argue that any such $\Phi$ could be decomposed into immediate reward, by defining the immediate reward of an action as the maximum achievable degree of $\Phi$ conditional on the taking of that action.  However, this requires $p_t(s_t,x_t)$ to involve not only straightforward data about the state $s_t$ and action $x_t$ but also data regarding the past history and possible futures following on from $s_t$.   That is, it fits into the DDS framework only if one assumes a setup in which each state $s_t$ contains very rich information about past and future history, and exists not as a delimited package of information about a system at one point in time, but rather as a sort of pointer to a certain position within a broader past/future history.   In this sort of setup one can define immediate rewards that depend on past and future history.   This sort of setup allows one to bypass certain conceptual pathologies associated with conventional RL and "wireheading" (wherein a system achieves its goals in a misleading way by modifying itself in such a way as to modify its interpretation of its goals) \cite{Goertzel2009RL}.   However, it also eliminates the simple practicality of the standard interpretations of the DDS / dynamic programming / RL framework, and one could argue eliminates most of the value of this framework.  

Another approach to generalize the standard dynamic programming formulae would be to stick with the localized definition of states and the more restricted sort of immediate reward -- but to accept that the actions a system takes may affect its long term memory, in such a way that after a system has chosen a certain action at a certain point in execution of a DDS process, the knowledge generated by this action might in some cases affect the result that would be obtained if it were to redo some of the immediate-reward calculations it has already made earlier while executing the DDS process.   In cases like this, one cannot guarantee that the decision ultimately made by the system while pursuing immediate reward step-by-step is going to result in maximization of the overall reward for the system, as judged by the system based on all the knowledge it has in its memory at the end of the DDS process.  

One could represent this sort of process schematically via assuming that the states $s_t$ contain knowledge that is mutable based on chosen actions, and looking at a modified functional equation of the form

\begin{align*}
f_t(s_t)=\max_{x_t\in X_t(s_t)} \biggl\{ (\text{expected reward during stage } t\mid s_t,x_t) + \\
 \alpha\sum_{s_{t+1}} \Pr(s_{t+1}\mid s_t,x_t)f_{t+1}(s_{t+1}) + \\
 \alpha\sum_{s_{t-1}} \Pr(s_{t-1}\mid s_t,x_t)f_{t-1}(s_{t-1})  
  \biggr\}
\end{align*}

\noindent where $\Pr(s_{t-1}\mid s_t,x_t)$ denotes the probability of $s_{t-1}$ being the state that leads up to $s_t$ calculated across the  possible worlds where  $s_{t-1}$ references a knowledge base incorporating the changes that were made in the course of choosing action $x_t$ during $s_t$.   However this sort of modified time-symmetric Bellman equation lacks an elegant iterative solution like the ordinary Bellman equation.

The practical approach we suggest for modeling currently relevant real-world cognitive systems is to take the ordinary expected reward based DDS as formulated above, with states and immediate rewards defined in a temporally localized way as usual, as a first approximation to cognitive dynamics.   It is not a fully accurate model of cognitive dynamics because states do sometimes encapsulate future and past histories in subtle ways, and because actions impact knowledge in ways that impact systems' retrospective views of previously calculated immediate rewards.   However it's a reasonable model of a lot of cognitive dynamics.   Large deviations from this approximation are cognitively interesting and sometimes important.  Small deviations may be modeled by stochastic variation on conventional dynamic programming, though we do not pursue this in depth here.

In particular, we suggest, deliberative abstract cognition is sometimes able to deviate from the simplistic DDS approach and incorporate a more holistic view of action selection incorporating past and future history.   This may be important for the dynamics of top-level goals in deliberative cognitive systems, and e.g. for the ability of advanced cognitive systems to avoid wireheading traps that expected reward maximizers will more easily fall into.   On the other hand, we hypothesize that effective large-scale cognition using limited computational resources generally depends on setting up situations where a more straightforward greedy or dynamic programming like DDS approach applies.

\subsection{Combinatory-Operation-Based Function Optimization}

Now we consider a particular sort of computational challenge, involving the making of a series of decisions involving how to best use a set of combinational operators $C_i$ to gain information about maximizing a function $F$ (or Pareto optimizing a set of functions $\{F_i\}$) via sampling evaluations of $F$ ($\{F_i\}$).   For simplicity we'll present this process -- which we call COFO for Combinatory-Operation Based Function Optimization -- in the case of a single function $F$ but the same constructs work for the multiobjective case.   We will show here how COFO can be represented as a Discrete Decision Process, which can then be enacted in greedy or dynamic programming style.

Given a function $F: X \rightarrow R$ (where $X$ is any space with a probability measure on it and $R$ is the reals), let $\mathcal{D}$ denote a "dataset" comprising finite subset of the graph $\mathcal{G}(F)$ of $F$, i.e. a set of pairs $(x,F(x))$.   We want to introduce a measure $q_F(\mathcal{D})$ which measures how much guidance $\mathcal{D}$ gives toward the goal of finding $x$ that make $F(x)$ large.   

The best measure will often be application-specific.  However, it is also worth articulating a sort of "default" general-purpose approach.   One way to do this is to assume a prior probability measure over the space $\mathcal{F}$ of all  measurable functions $f$ with the same domain and range as $F$.   Given a threshold value $\rho$ and a function $f \in \mathcal{F}$, let $M_\rho(f)$ denote the set of $x$ so that $f(x) \geq f(y)$ with probability at least $1 - \rho$ for a randomly selected $y \in X$.  I.e. $M_\rho(f)$ is the set of arguments $x$ for which $f(x)$ is in the top $\rho$ values for $f$.   One can then set $q_{\rho,F}(\mathcal{D})$ equal to the entropy of the set $M_\rho^D = \bigcup_{f: D \in \mathcal{G}(f)}   M_\rho(f)$.

If this entropy is large, then knowing $f$ agrees with $F$ on $D$ doesn't narrow things down much in terms of finding $x$ that make $F$ big -- it still leave a wide spread of possibilities.   If this entropy is small, then the set of $x$ that make $f$ big for functions $f$ that agree with $F$ on $D$ is a strongly localized distribution in  $\mathcal{F}$ -- so that knowing $D$ meaningfully focuses the search space for $x$ that make $F$ big.

Given two such datasets  $D_1$ and $D_2$ one can of course look at the relative quality $q_{\rho,F}(D_1, D_2 ) $ equal to the relative entropy of $M_\rho^{D_1}$ given $M_\rho^{D_2}$.

We can then look at greedy or dynamic programming processes aimed at gradually building a set $D$ in a way that will maximize  $q_{\rho,F}(D)$.   Specifically, in a cognitive algorithmics context it is interesting to look at processes involving combinatory operations $C_i: X \times X \rightarrow X$ with the property that 

$$
P( C_i(x, y  ) \in M_\rho^D | x \in M_\rho^D, y \in M_\rho^D)  \gg P(z \in M_\rho^D | z \in X)
$$

\noindent. That is, given $x, y \in M_\rho^D$, combining $x$ and $y$ using $C_i$ has surprisingly high probability of yielding $ z \in M_\rho^D$.

Given combinatory operators of this nature, one can then approach gradually building a set $D$ in a way that will maximize  $q_{\rho,F}(D)$, via a route of successively applying combinatory operators $C_i$ to the members of a set $D_j$ to obtain a set $D_{j+1}$.

Framing this COFO process in terms of the above-described Discrete Decision Process:

\begin{enumerate}
\item A {\bf state} $s_t$ is a dataset $D$ formed from function $F$
\item  An {\bf action} is the formation of a new entity $z$ by 
\begin{enumerate}
\item Sampling $x, y$ from $X$ and $C_i$ from the set of available combinatory operators,  in a manner that is estimated likely to yield $z=C_i(x,y)$  with $z \in M_\rho^D$
\begin{enumerate}
\item As a complement or alternative to directly sampling, one can perform probabilistic inference of various sorts to find promising $(x,y,C_i)$.   This probabilistic inference process itself may be represented as a COFO process, as we show below via expressing PLN forward and backward chaining in terms of COFO
\end{enumerate}
\item Evaluating $F(z)$, and setting $D^* = D \cup (z, F(z))$.
\end{enumerate}
\item  The {\bf immediate reward} is an appropriate measure of the amount of new information about making $F$ big that was gained by the evaluation $F(z)$.   The right measure may depend on the specific COFO application; one fairly generic choice would be the relative entropy $q_{\rho,F}(D^*, D)$ 
\item  {\bf State transition}: setting the new state $s_{t+1}=D^*$
\end{enumerate}

\noindent A concurrent-processing version of this would replace 2a with a similar step in which multiple pairs $(x,y)$ are concurrently chosen and then evaluated.

In the case where one pursues COFO via dynamic programming, it becomes {\it stochastic} dynamic programming because of the probabilistic sampling in the action.   The sampling step in the above can be specified in various ways, and incorporates the familiar (and familiarly tricky) exploration/exploitation tradeoff.   If probabilistic inference is used along with sampling, then one may have a peculiar sort  of stochastic dynamic programming in which the step of choosing an action involves making an estimation that itself may be usefully carried out by stochastic dynamic programming (but with a different objective function than the objective function for whose optimization the action is being chosen).

Basically, in the COFO framework we  are looking at the process of optimizing $F$ as an explicit dynamical decision process conducted via sequential application of an operation in which: Operations $C_i$ that combine inputs chosen from a distribution induced by prior objective function evaluations, are used to get new candidate arguments to feed to $F$ for evaluation.   The reward function guiding this exploration is the quest for reduction of the entropy of the set of guesses at arguments that look promising to make $F$ near-optimal based on the evaluations made so far.

As noted at the start, the same COFO process can be applied equally well the case of Pareto-optimizing a set of objective functions.   The definition of $M_\rho^D$ must be modified accordingly and then the rest follows.

Actually carrying out an explicit stochastic dynamic programming algorithm according to the lines described above, will prove computationally intractable in most realistic cases.   However, we shall see below that the formulation of the COFO process as dynamic programming (or simpler greedy sequential choice based optimization) provides a valuable foundation for theoretical analysis.   

\section{Specific Cognitive Algorithms as Greedy or Dynamic Programming Based Decision Processes on Metagraphs}

The next thread in the tapestry we're weaving here is the observation that the cognitive algorithms centrally used in the OpenCog AGI architecture, which encompass well known AI paradigms like logical reasoning, evolutionary learning and clustering -- can be formulated in COFO terms, where the functions $F$ involved are defined on subgraphs of typed metagraphs (TMGs).   This allows us to argue that these AI algorithms can be viewed as judicious, computationally practical approximations to greedy optimization or stochastic dynamic programming for COFO in a TMG context -- somewhat as classic RL is an effective approximation to dynamic programming in other cases.   

Looking forward to Section \ref{sec:Galois} below, this will then allow us to represent these cognitive operations as Galois connections, thus enabling us to represent their solutions elegantly in terms of hylomorphisms and chronomorphisms on TMGs.   The role of dynamic programming here is then, in part, to serve as the connecting formalism between a diverse set of cognitive algorithms and a general formulation in terms of chronomorphisms.

Given that standard RL is also an effective approximation of stochastic dynamic programming for different sorts of states and actions, it seems this formal approach may also provide an effective way to think about the integration of RL with other cognitive algorithms such as logical inference and evolutionary learning.   I.e. if  one considers states that include the sorts of real vectors that conventional RL works well with, along with the sub-TMGs that various other cognitive algorithms work well with, then one can think about an approximate approach to solving the corresponding dynamic programming problem that incorporates common RL-type model learning mechanisms with combinatory-operation-based learning mechanisms as are embodied in these other cognitive algorithms.

We will focus here on the algorithmic processes involved in the various cognitive functions considered, rather than on knowledge representation issues, but this is because we are assuming the context of the OpenCog framework \cite{EGI1} \cite{EGI2}, in which the various cognitive algorithms considered represent their input and output data and their internal states and structures in terms of the optionally-directed typed metagraph called the Atomspace.   We also assume that the type system on the Atomspace metagraph includes in its vocabulary the paraconsistent and probabilistic types described in \cite{goertzel2021paraconsistent}.   The core actions involved in most of the cognitive algorithms described here involve sampling from distributions on a metagraph -- i.e. "metagraph probabilistic programming" as described in \cite{goertzel2021paraconsistent}.   What we are doing here is leveraging the process of sampling from distributions on subgraphs of typed metagraph to create cognitive algorithms, via observing that a great variety of important cognitive algorithms can be represented in terms of a fairly stereotyped discrete decision process involving pursuit of function optimization via iterative application of combinatory operators acting on sampled sub-metagraphs.

\subsection{Forward Chaining Inference}

Forward chaining inference in Probabilistic Logic Networks \cite{PLN} or other similar frameworks may be based on a quality metric measuring the interestingness or surprisingness of an inference conclusion (see \cite{PatternMining} for a formalization of surprisingness in an OpenCog context).   The general process involved is then, qualitatively: 

\begin{itemize}
\item Choose reasonably interesting premises and combine them with an inference rule to get a conclusion that is hopefully also interesting.   
\item Select the inference rule based on past experience regarding which rules tend to work for generating interesting conclusions from those sorts of premises.
\end{itemize}

\noindent In this case the objective function $F$ of the COFO process is the interestingness measure.

The use of probabilistic inference for choosing actions within the action-selection step of a stochastic dynamic programming approach to COFO may be formulated in this way. Here one is seeking $(x,y,C_i)$ so that $\chi_{M_\rho^D}(C_i(x,y))$ is large, where $\chi_{M_\rho^D}(z)$ is the fuzzy membership degree of $M_\rho^D$ considered as a fuzzy set.   So one has $F(x,y,C_i) = \chi_{M_\rho^D}(C_i(x,y))$, which in context is a measure of the interestingness of $(x,y,C_i)$.

We then have a DDS formalization via

\begin{enumerate}
\item A {\bf state} $s_t$ is a set of statements labeled with interestingness values
\item  An {\bf action} is the formation of a new statement by 
\begin{enumerate}
\item Sampling premise statements $x, y$ from the knowledge base and $C_i$ from the set of available inference rules,  in a manner that is estimated likely to yield $z=C_i(x,y)$  that is interesting
\begin{enumerate}
\item As a complement or alternative to directly sampling, one can perform probabilistic inference to find promising premise / inference-rule combinations in context.   This is "inference-based inference control."  
\end{enumerate}
\item Doing the selected inference to create a statement $z$, evaluating this statement's interestingness and creating an updated state by adding the statement labeled with its interestingness to the previous state
\end{enumerate}
\item  The {\bf immediate reward} is the amount of new information gained about how to find interesting statements, via the process of finding $z$ and evaluating its interestingness
\item  {\bf State transition}: setting the new state equal to the updated state. just created, and then iterating again by returning to "Sampling premise statements..." etc.
\end{enumerate}

\noindent A concurrent version of this would sample multiple $(x,y,C_i)$ concurrently in step 2a.   Similar strategies work for making all the cognitive processes considered below concurrent, so we will not explicitly note this in each case.

The key conceptual constraint placed on the measure of interestingness used here is that it must be meaningful to assess the interestingness of an inference dag as the sum of the interestingnesses of the nodes in the dag.   For instance one could define the interestingness of an inference result via mutual information.  A probabilistic logical statement $S(v)$, with variable-arguments $v$ drawn from a knowledge base (such as an OpenCog Atomspace), can be associated with a probability distribution $\nu$ over the knowledge-base in which $\nu(v)$ is proportional to $S(v)$.   

One can then calculate entropy (information) and mutual information among such distributions.   Interestingness of an inference result can then be measured as the conditional information-theoretic surprisingness (information gain) of the result, where the conditioning is done on a combination of the predecessors of the inference in the inference dag plus the assumed background knowledge.   The needed linear decomposition then emerges from the algebra of mutual information, if the knowledge base is assumed constant throughout the inference.   

Alternatively, one could define a distinction graph \cite{Graphtropy} on the knowledge base via drawing a link between $v$ and $w$ if $|S(v) - S(w)| < \epsilon$, and then calculate absolute and conditional graphtropies of these distinction graphs.    Interestingness of an inference result can then be measured as the conditional graphtropy of the result, where the conditioning is done on a combination of the predecessors of the inference in the inference dag plus the assumed background knowledge.   The needed linear decomposition follow from graphtropy algebra, if the knowledge base is assumed constant throughout the inference.

A more efficient alternative to trying to explicitly solve the stochastic dynamic programming functional equation, in this case, is to build a large model of which series of choices $(x,y,C_i)$ seem to be effective in which contexts.   This is what PLN "adaptive inference control" aims to do \cite{PLN}.   Doing explicit stochastic dynamic programming but using the (2ai) "inference based inference control" option is one way of implementing this, as each instance of the underlying inference can make use of knowledge generated and saved when applying similar inference elsewhere in the dynamic programming process.

\subsection{Backward Chaining Inference}

To give a similar analysis for uncertain backward chaining, two kinds of BC inference need to be considered: TV inference, aimed at estimating the truth value of a statement; and SS inference, aimed at finding some entity that satisfies a predicate (lies in the SatisfyingSet of the predicate).   

To formalize backward chaining TV inference as stochastic dynamic programming, we start with the idea of a "backward inference dag" (BID) define as a binary dag whose internal nodes $N$ are labeled with $(N_s, N_r)$ = (statement, inference rule) pairs.  A leaf node may either be similar to an internal node, or may have a label of the form $(N_s,N_D)$ where $D$ is a dataset on which the truth value of $N_s$ may be directly estimated.   A BID encapsulates a set of inferences leading from a set of premises (the leaves of the tree) to a conclusion (the root of the tree).   This is uncertain inference so the statements involved are labeled with truth values.  The semantics is that the statement at a node $N$ is derived from the statements at its children, using the rule at $N$.     Leaf nodes that aren't labeled with datasets have empty inference rules in their labels; internal nodes cannot have empty inference rules.   

To formalize backward chaining SS inference as stochastic dynamic programming, we use a slightly different BID whose internal nodes $N$ are labeled with $(N_P, N_r, N_g)$ tuples.   Here $N_P$ is a predicate, $N_r$ is an inference rule, $N_g$ is a goal qualifier.    As in the TV case, leaf nodes have empty inference rules in their labels; internal nodes cannot.   The goal qualifier may e.g. specify that the goal of the inference is to work toward finding some entity that makes the predicate $N_P$ true, or rather to work toward finding some entity that makes $N_P$ false.  Leaf nodes are of the form $(N_P,N_E, N_g)$ where $E$ is some specific entity that fulfills the predicate $N_P$  according to $N_g$ (to an uncertain degree encapsulated in the associated truth value).

We will assume here use of PLN Simple Truth Values (STV) including strength (probability) values $s$ and confidence values $c$ (representing "amount of evidence" scaled into $[0,1]$.    To quantify the amount of reward achieved in expanding the BID and thus moving from a prior truth value estimate $(s_1,c_1)$ to a new truth value estimate $(s_2,c_2)$ for the root statement, we introduce the CWI (Confidence Weighted Information Gain) measure.   Referring to the fundamental theory of STVs as outlined in \cite{PLN}, let $\gamma_{(s,c)}$ denote the  second-order distribution corresponding to the pair $(s,c)$.   We then define the CWIG as the information gain in the distribution $\gamma_{(s_2,c_2)}$ conditioned on $\gamma_{(s_1,c_1)}$ plus background knowledge; we can also make similar calculations using graphtropy, yielding a Confidence Weighted Graphtropy Gain (CWGG).

For TV inference we then have

\begin{enumerate}
\item A {\bf state} $s_t$ is a BID (which may be viewed from a COFO perspective as a set of sub-BIDS paired with CWIG/CWGG values)
\item  An {\bf action} is the formation of a new statement by 
\begin{enumerate}
\item Sampling a leaf node $N$ from the current BID, along with premises $x, y$ from the knowledge base and a rule $C_i$ from the set of available inference rules,  in a manner that is estimated likely to yield $N_s=C_i(x,y)$  that has a high truth value confidence
\begin{enumerate}
\item As a complement or alternative to directly sampling, one can perform probabilistic inference to find promising premise / inference-rule combinations in context.   This is "inference-based inference control."  
\end{enumerate}
\item Doing the selected inference to estimate the truth value of $N_s$ and then updating the BID by adding children to $N$ corresponding to $x$ and $y$
\end{enumerate}
\item  The {\bf immediate reward} is the CWIG (or CWGG) achieved regarding the truth value of the statement at the root node of the BID, via the process of estimating the truth value of $N_s$ via  $N_s=C_i(x,y)$
\item  {\bf State transition}: setting the new BID equal to the updated BID
\end{enumerate}

For SS inference, the story is similar:

\begin{enumerate}
\item A {\bf state} $s_t$ is a BID 
\item  An {\bf action} is the formation of a new statement by 
\begin{enumerate}
\item Sampling a leaf node $N$ from the current BID, along with premise predicates $x, y$ from the knowledge base and a rule $C_i$ from the set of available inference rules,  in a manner that is estimated likely to yield $N_P=C_i(x,y)$  so there is a high $s*c$ to the implication that if some entity $E$ fulfills predicates $x$ and $y$ it also fulfills $P$.
\begin{enumerate}
\item As a complement or alternative to directly sampling, one can perform probabilistic inference to find promising premise / inference-rule combinations in context.   This is "inference-based inference control."  
\end{enumerate}
\item Updating the BID by adding children to $N$ corresponding to $x$ and $y$
\end{enumerate}
\item  The {\bf immediate reward} is the CWIG (or CWGG) gained regarding the satisfaction of the predicate at the root of the BID by some entity, via the process of extending the BID with the inference of $N_P$ from  $N_P=C_i(x,y)$
\item  {\bf State transition}: setting the new BID equal to the updated BID
\end{enumerate}

\subsection{Evolutionary Learning}

Evolutionary optimization, in a fairly general form, can be straightforwardly cast in terms of COFO and greedy optimization.  Here we consider an evolutionary approach to finding $x$ that maximize, or come close to maximizing, the "fitness function" or objective function $F: X \rightarrow R$:

\begin{enumerate}
\item A {\bf state} $s_t$ is a population of candidate solutions (genotypes)
\item  An {\bf action} is the addition of a new member to the population via one of the operations
\begin{enumerate}
\item Sampling in the form of either
\begin{enumerate}
\item Sampling pairs  $(x, y)$ of genotypes from the population and $C_i$ from the set of available crossover operators,  in a manner that is estimated likely to yield $z=C_i(x,y)$  that is fit
\item Sampling an individual genotype $x$ from the population and $C_i$ from the set of available mutation operators,  in a manner that is estimated likely to yield $z=C_i(x)$  that is fit
\end{enumerate}
\item As a complement or alternative to directly sampling, one can perform probabilistic inference to find promising genotypes and crossover or mutation operators or combinations thereof in context.   Or one can bypass these operators  and choose a new genotype $z$ using a generative model learned from the set of relatively highly fit population members.   The latter gives one an "Estimation of Distribution Algorithm" \cite{Pelikan2002}.
\end{enumerate}
\item Enacting the selected approach to create a new genotype $z$, evaluating this genotype's fitness and creating an updated population by adding the new element to the population
\item  The {\bf immediate reward} is the amount of new information gained about how to find large values of $F$, via the process of finding $z$ and evaluating its fitness
\item  {\bf State transition}: setting the new population equal to the prior population.
\end{enumerate}

Ordinarily evolutionary algorithms follow a single "greedy" path from an initial population, growing the population till it comes to contain a good enough solution.   An explicit dynamic programming approach would make sense only in a situation where, say, the fitness function was constantly changing so that one had to keep re-evaluating the population members -- in which case one would have a pressure militating in favor of a smaller population, so that sculpting an optimal population rather than simply growing the population endlessly would make sense.   Of course the dynamic fitness function scenario has resonance with the properties of many real ecological niches -- in real evolutionary ecology, the fitness function does change as the environment changes, and population size is restricted by resource limitations.

It is interesting to think about variations on evolutionary algorithms that relate to dynamic programming in roughly the same way standard RL algorithms do.   I.e., one can update, during an evolutionary optimization process, a probabilistic model of "what is a good population" for exploring the given function's graph with a view toward optimization -- and then grow a population, using mutation and crossover and EDA type modeling, in a direction according to this probabilistic model.   This is a way of implicitly doing what the dynamic programming approach is doing but without the massive time overhead.  

A concrete algorithm along these lines is the Info-Evo approach suggested in \cite{Goertzel2021evo}.   In this approach one looks at the space of weighting distributions over the elements of an evolving population,  estimates the natural gradient in this space, and then at each stage seeks to evolve the population along the direction of the natural gradient.   In terms of the above flow, this corresponds to Option 2b, where the probabilistic inference used to find promising genotypes and operators involves estimating the natural gradient on population-weighting-distribution space and then preferentially choosing genotypes and operators that are likely to move the population as a whole along the natural gradient.

OpenCog's MOSES algorithm adds an extra layer to evolutionary optimization.   One defines a "deme" as a set of program dags, and does evolution by mutation and crossover and probabilistic modeling within each deme.    Then there is a deme level of learning, in which poorly performing demes are killed and highly performing demes are copied.

\begin{enumerate}
\item A {\bf state} $s_t$ is a meta-population of demes
\item  An {\bf action} is one of
\begin{enumerate}
\item Sampling in the form of either
\begin{enumerate}
\item Selecting a deme from the meta-population that appears to have a high probability of failure, and removing it
\item Selecting a deme from the meta-population that appears to have a high probability of success, and cloning it
\item Selecting a deme from a uniform distribution on the the meta-population and advancing its growth by carrying out a sequence of actions within its own evolutionary process
\end{enumerate}
\item As a complement or alternative to directly sampling, one can perform probabilistic inference to find promising or unpromising demes for appropriate action.   This gives one an EDA on the deme level, which has not yet been experimented with.   
\end{enumerate}
\item Updating the deme population, or the modified deme, accordingly
\item  The {\bf immediate reward} is the amount of new information gained about how to find large values of $F$, via the process of curating or updating demes
\item  {\bf State transition}: setting the new deme population equal to the prior deme population.
\end{enumerate}

\subsection{Nonlinear-Dynamical Attention Allocation}

Economic Attention Allocation (ECAN) is the process in OpenCog that spreads Short-Term Importance (STI) and Long-Term Importance (LTI) values among Atoms in the Atomspace.   The simplest way to view the use of these values is: STI values guide which Atoms get processing power, LTI values guide which Atoms remain in RAM.

STI corresponds conceptually to a DDS such as:

\begin{enumerate}
\item A {\bf state} $s_t$ is a set of Atoms labeled with short-term importance (STI) values
\item  An {\bf action} is the process of 
\begin{enumerate}
\item Sampling an Atoms $x$ from the knowledge base and another Atom $y$ connected to $x$ via some Atomspace link
\item Subtracting $Q$ from $x$'s STI and adding $Q$ to $y$'s STI
\end{enumerate}
\item  The {\bf immediate reward} is the amount of utility obtained by the overall system, during the immediately future interval, that is causally attributable to $x$ and $y$ (via whatever causal attribution mechanisms are operative).   
\item  {\bf State transition}: setting the new state equal to the updated state just created, with the new STI values for $x$ and $y$
\end{enumerate}

The story for LTI is basically the same, the difference between the two cases being wrapped up in the causal attribution estimate used to calculate immediate reward.   In the case of LTI we assume the background activity of a "forgetting" process that removes Atoms with overly low LTI from the Atomspace.

To make this work in terms of expected reward maximization, we need the measure of causality to be linearly decomposable, so that the amount of causal influence attributed to a causal chain is equal to the sum of the causal influences attributed to the elements in the chain.  This generally works with measures of causality associated with Judea Pearl style causal networks \cite{pearl2009causality} and also with treatment of causation as conditional implication plus temporal precedence as discussed in \cite{PLN}.   If one treats causation as conditional implication plus temporal predecence plus existence of a plausible causal mechanism (as also discussed in \cite{PLN}), then one needs to take care that one's methods of assessing existence of plausible causal mechanism adhere to decomposability; e.g. if one uses uncertain logic to assess plausiblity of causal mechanisms then this can be made to work along the lines of the treatment of inference processes described above.

In this case, actually doing attention allocation using dynamic programing would be insanely expensive, but would allow one to figure out an optimal path for attention allocation in a given Atomspace.  What is actually done in OpenCog systems is to implement a fairly simple activation-spreading-based method for diffusing STI and LTI through the Atomspace.  This activation-spreading method is intended as a crude but cheap approximation of the attention allocation path that dynamic programming would give.

\subsection{Agglomerative Clustering}

Agglomerative clustering follows a similar logic to forward-chaining inference as considered above -- cluster quality plays the role of interestingness, and the application of a merge operator takes the place of application of an inference rule.

\begin{enumerate}
\item A {\bf state} $s_t$ is a "clustering", meaning set of Atoms grouped into clusters
\item  An {\bf action} is the formation of a new clustering
\begin{enumerate}
\item Sampling clusters $x, y$ from the current clustering and $C_i$ from the set of available cluster-merge rules,  in a manner that is estimated likely to yield $z=C_i(x,y)$  that increases clustering quality
\begin{enumerate}
\item As a complement or alternative to directly sampling, one can perform probabilistic inference to find promising cluster combinations in context.   This is "inference-based clustering control."  
\end{enumerate}
\item Performing the selected merger to create a new cluster $z$, evaluating the clustering quality after replacing $x,y$ with $z$ and creating an updated overall clustering
\end{enumerate}
\item  The {\bf immediate reward} is the increase in clustering quality obtained via the agglomeration of $x$ and $y$ into $z$
\item  {\bf State transition}: setting the new clustering equal to the updated clustering
\end{enumerate}

\subsection{Pattern Mining}

Pattern mining follows a similar template to evolutionary learning, with crossover/mutation operations replaced by operations for expanding known patterns by adding new terms to them.

\begin{enumerate}
\item A {\bf state} $s_t$ is a population of patterns mined from an Atomspace
\item  An {\bf action} is the addition of a new pattern to the population via 
\begin{enumerate}
\item selecting a pattern $x$ from the population, with a probability proportion to its pattern-quality, then selecting a pattern-expansion operator $C_i$ with a probability proportional to the estimated odds that applying it to $x$ will yield a high-quality pattern
\item applying $C_i$ to $x$ to obtain a new candidate pattern $z=C_i(x)$
\item Evaluating the new pattern $z$'s pattern-quality
\end{enumerate}
\item  The {\bf immediate reward} is the amount the total pattern-quality of the set of patterns in the population has increased, as a result of adding $z$ to the population
\item  {\bf State transition}: adding $z$ to the pattern population
\end{enumerate}

\noindent As in the evolutionary learning case, here the default approach is to proceed in a greedy way, but approximations to a richer stochastic dynamic programming strategy could also be interesting.

\section{Alignment of Subpattern Hierarchies with Optimization Subgoal Hierarchies}

There is an interesting correspondence between the subgoal hierarchies involved in the dynamic-programming-type DDSs described above, and the subpattern hierarchies described in \cite{goertzel2020grounding}.   In the latter paper it is shown that, if one has a collection of mutually associative combinatory operations, then one can use these to describe a subpattern hierarchy, i.e. a dag in which $x$ is a child of $y$ if there is some $z$ and some combinatory operator $C$ so that

\begin{itemize}
\item $C(y,z) = x$
\item $\sigma(y) + \sigma(z) + \sigma^*(C,y,z) < \sigma(x)$
\end{itemize}

\noindent where $(\sigma, \sigma^*)$ is a simplicity measure.

Formally speaking, then, if one has a DDS based on a set of combinatory operators $C_i$ that are mutually associative, there is a subpattern hierarchy connected with that DDS.   There are also alternate routes to getting subpattern hierarchies out of combinatory operator collections that don't require mutual associativity, and these may relate to some sorts of DDSs also, but the mutually-associative case is the most relevant one here.

Just because a dynamic-programming-based DDS formally corresponds to a subpattern hierarchy, doesn't mean this correspondence necessarily has to be meaningful in terms of the operation of the DDS.   However, it happens that for cognitively interesting DDS's there often is a meaningful correspondence of this nature.   This sort of correspondence might be described as a cognitive synergy between perception and action -- or put more precisely, between pattern-recognition and action-selection.

\subsection{Inferential Subpatterns, Reversible Logic and Reversible Computing}

In the case of DDSs embodying uncertain inference for example, the relevant simplicity measure is the {\it count}, i.e. the number of items of evidence on which a given judgment or hypothesis relies.   Suppose $C$ is an inference rule used to combine premises $x$ and $y$ to obtain some conclusion.  If $x$ and $y$ are based on disjoint bodies of evidence, then the count of $C(y,z)$ can be effectively defined as the sum of the count of $y$ and the count of $z$.   This is an upper bound for the count of  $C(y,z)$ in the general case.

Are the multiple inference rules $C_i$ involved in a logic system like PLN mutually associative?  This turns out to depend sensitively on how the rules are formulated.   In most logics, ordinary one-way (conditional) implication is not associative, but  two-way (biconditional) implication is associative.  So if one formulates one's inference rules as equivalences rather than one-way implications, then -- at least within the domain of crisp inference rules, setting aside quantitative uncertainty management -- one obtains a formulation of logic in terms of mutually associative rules.  Another way to say this is: A reversible formulation of logic will tend to involve mutually associative inference rules.

Reversible forms of Boolean logic gates are well known \cite{pan2005reversible} and play interesting roles in quantum computing \cite{peres1985reversible}.   Reversible forms of predicate logic have also been fleshed out in detail, e.g.  Sparks and Sabry's Superstructural Reversible Logic \cite{sparks2014superstructural}, which -- just as linear logic adds a second version of' "and"-- adds a second version of ?or?.  The second "or operator is used to ensure the conservation of choice (or -- equivalently  -- the conservation of temporal resources) just as linear logic ensures the conservation of spatial resources.   There would appear to be an isomorphic mapping between this logic and Constructible Duality logic with its new exponentials that form a mirror image of the standard linear logic exponentials \cite{patterson1998implicit}.

Reversible logic  maps via a Curry-Howard type correspondence into reversible computing, via the observation that every reversible-logic proof corresponds to a computation witnessing a type isomorphism.   One thus concludes that, if one wants an inference process that corresponds naturally to a subpattern hierarchy, one wants to use reversible inference rules, which then turns one's inference process into a process of executing a reversible computer program.

The elegance of the achievement of mutually associative inference operations via use of co-implication somewhat recedes when one incorporates uncertainty into the picture.   $A \xleftrightarrow{\nu_1} B \xleftrightarrow{\nu_1}  C \xleftrightarrow{\nu_1}  D$, where the $\nu_i$ are probability distributions characterizing the co-implications, may indeed give different results depending on how the co-implications are parenthetically grouped, related to the probabilistic dependencies between the distributions.   If all the distributions are mutually independent then associativity will hold, but obviously this won't always be the case.   One can try to achieve as much independence as possible, and one can also try to arrange dependencies in an order copacetic with the processing one actually needs to do, a topic to be revisited below.

\subsection{Subpattern Hierarchy in Agglomerative Clustering and Pattern Mining}

The coordination of subgoal hierarchy with subpattern hierarchy can also be seen in various other cognitive processes, including some considered above.

In the context of agglomerative clustering, one may define the simplicity of a clustering in information-theoretic terms, e.g. as the logical entropy of the partition constituted by the clustering.   Looking at cluster merge operations that act to map of a clustering into a less fine-grained clustering via joining certain clusters together, it's obvious that the merge operations are associative (they are just unioning disjoint subsets, and union is associative).   

For pattern mining, relevant simplicity measures would be the frequency of a pattern or the information-theoretic surprisingness of a pattern.   Pattern combination is associative.  For instance, if patterns are represented as metagraphs (which is the case in OpenCog's Pattern Miner) then we can combine two patterns with metagraph Heyting algebra conjunction and disjunction operations, in which case associativity follows via the rules of Heyting algebra.   

The process of pattern growth is then representable as a matter of taking patterns that are both small in size (as metagraphs) and simple (by the simplicity measure), and combining them conjunctively and disjunctively to form new patterns which are then pruned based on their simplicity -- and then iteratively repeating this process.   One can then spell out algebraic relationships and inequalities regarding the frequency or surprisingness of the conjunction or disjunction of two patterns, providing a connection between the subgoal hierarchy used in pattern growth during the pattern mining process, and the subpattern hierarchy implied by the simplicity measure.

\section{Cognitive Algorithms as Galois Connections Representing Search and Optimization on Typed Metagraphs}
\label{ref:Galois}

One interesting aspect of the formulation of multiple cognitive algorithms as DDSs embodying COFO processes is the guidance it gives regarding practical implementation of these cognitive algorithms.   Following the ideas on Galois connection based program derivation outlined in \cite{mu2012programming}, we will trace here a path from DDS formulation of cognitive algorithms to representation of these cognitive algorithms in terms of folding and unfolding operations.   

First, as a teaser, we will leverage the Greedy Theorem from \cite{mu2012programming} to show that greedy steepest-ascent (or steepest-descent) optimization is represented naturally as a fold operation.  Then, proceeding to dynamic programming type algorithms, we will show that if one has a DDS embodying COFO based on mutually associative combinatory operations, it can be implemented as a hylomorphism (fold followed by unfold) on metagraphs or, if one wants to take advantage of memory caching to avoid repetitive computation, a metagraph chronomorphism (history-bearing fold followed by history-bearing unfold).   This suggests that one strategy for efficient implementation of cognitive algorithms is to create a language in which (pausable and resumable) metagraph folds and unfolds of various sorts are as efficient as possible in both concurrent and distributed processing settings -- an important conclusion for the design of next-generation AGI systems like OpenCog Hyperon \cite{goertzel2020kind}.

\subsection{Greedy Optimization as Folding}

Suppose we are concerned with maximizing a function $f:X \rightarrow R$  via a ``pattern search" approach.  That is, we assume an algorithm that repeatedly iterates a pattern search operation such as: Generates a set of candidate next-steps from its focus point $a$, evaluates the candidates, and then using the results of this evaluation, chooses a new focus point $a^*$.   Steepest ascent obviously has this format, but so do a variety of derivative-free optimization methods as reviewed e.g. in \cite{torczon1995pattern}.

Evolutionary optimization may be put in this framework if one shifts attention to a population-level function $f_P: X^N \rightarrow R$ where $X^N$ is a population of $N$ elements of $X$, and defines $f_P(x)$ for $x \in X^N$ as e.g. the average of $f(x)$ across $x \in X^N$ (so the average population fitness, in genetic algorithm terms).   The focus point $a$ is a population, which evolves into a new population $a^*$ via crossover or mutation -- a process that is then ongoingly iterated as outlined above.

The basic ideas to be presented here work for most any topological space $X$ but we are most interested in the case where $X$ is a metagraph.  In this case the pattern search iteration can be understood as a walk across the metagraph, moving from some initial position in the graph to another position, then another one, etc.

We can analyze this sort of optimization algorithm via the Greedy Theorem from \cite{mu2012programming}, 

\begin{thm}
(Theorem 1 from \cite{mu2012programming}). $\llparenthesis S \upharpoonright R \rrparenthesis \subseteq \llparenthesis S \rrparenthesis \upharpoonright R$ if R is transitive and S satisfies the "monotonicity condition" $R^\circ \leftarrow{S} FR^\circ$
\end{thm} \label{thm:Greedy}

\noindent which leverages a variety of idiosyncratic notation:

\begin{itemize}
\item $R \xleftarrow{S} FR$ \textrm{ indicates } $S \cdot FR \subseteq R \cdot S$
\item $\llparenthesis S \rrparenthesis$ \textrm{ means } the operation of folding $S$
\item $\langle \mu X :: f X \rangle$ denotes the least fixed point of $f$
\item $T^\circ$  \textrm{ means } the converse of $T$, i.e. $(b,a) \in R^\circ \equiv (a,c) \in R$
\item $S \upharpoonright R$  \textrm{ means } "$S$ shrunk by $R$", i.e. $S \cap R / S^\circ$
\end{itemize}

Here $S$ represents the local candidate-generation operation used in the pattern-search optimization algorithm, and $R$ represents the operatio of evaluating a candidate point in $X$ according to the objective function being optimized.   

The rhs of the equation in the Theorem describes a process in which the candidate-generation process is folded across all of $X$ and then all the candidates generated are evaluated.   I.e. this is basically an exhaustive search.   In the case where $X$ is a metagraph, we have in \cite{Goertzel2020metagraph} given a detailed formulation of what it means to fold across a metagraph, based on a model of metagraphs as composed of sub-metagraphs joined together via dangling-target-joining operations.  The exhaustive search is then a matter of folding candidate-generation across all the sub-metagraphs of the metagraph.   This is a valid fold operation so long as the set of candidates generated from the join of (say) three sub-metagraphs is independent of the order in which the joining occurs.

The lhs of the equation describes the folding across $X$ of a process in which: The candidate-generation process ($S$) is done locally at one location, and evaluation ($R$) is then carried out, resulting in a single candidate being selected.

The monotonicity condition, in this case, basically enforces that the objective function is convex.   In this case we lose nothing by computing local optima at each point and then folding this process across the space $X$ (e.g. the metagraph).   If the specifics of the pattern applies depend on the prior history of the optimization process then we have a histomorphism rather than just an ordinary catamorphism (fold), e.g. this is the case if there are optimization parameters that get adapted based on optimization history.

If the objective function is not convex, then the theorem does not hold, but the greedy pattern-search optimization may still be valuable in a heuristic sense.   This is the case, for instance, in nearly all real-world applications of evolutionary programming, steepest ascent or classical derivative-free optimization methods.

\subsection{Galois Connection Representations of Dynamic Programming Based DDSs Involving Mutually Associative Combinatory Operations} \label{sec:Galois}

Next we consider how to represent dynamic programming based execution of DDSs using folds and unfolds.  Here our approach is to leverage Theorem 2 in \cite{mu2012programming} which is stated as

\begin{thm}
(Theorem 2 from \cite{mu2012programming}). Assume $S$ is monotonic with respect to $R$, that is, $R \xleftarrow{S} F_R$ holds, and $dom(T) \subseteq dom(S \cdotp FM)$.   Then 

$$
M=(  \llparenthesis S \rrparenthesis \cdotp  \llparenthesis T \rrparenthesis^\circ ) \upharpoonright R \Rightarrow \langle \mu X :: (S \cdotp FX \cdotp T^\circ) \upharpoonright  R \rangle \subseteq M
$$

\end{thm} \label{thm:DP}

Conceptually, $T^\circ$ transforms input into subproblems, e.g.

\begin{itemize}
\item for backward chaining inference, chooses $(x,y,C)$ so that $z = C(x,y)$ has high quality (e.g. CWIG)
\item for forward chaining, chooses x, y, C so that z = C(x,y) has high interestingness (e.g. CWIG)
\end{itemize}

\noindent $FX$ figures out recursively which combinations give maximum immediate reward according to the relevant measure.   These optimal solutions are combined and then the best one is picked by $\upharpoonright R$, which is the evaluation on the objective function.   Caching results to avoid overlap may be important here in practice (and is what will give us histomorphisms and futumorphisms instead of simple folds and unfolds).

The fix-point based recursion/iteration specified by the theorem can of course be approximatively rather than precisely solved -- and doing this approximation via statistical sampling yields stochastic dynamic programming.  Roughly speaking the approach symbolized by $M=(  \llparenthesis S \rrparenthesis \cdotp  \llparenthesis T \rrparenthesis^\circ ) \upharpoonright R $  begins by applying all the combinatory operations to achieve a large body of combinations-of-combinations-of-combinations-$\ldots$, and then shrinks this via the process of optimality evaluation.  On the other hand, the least-fixed-point version on the rhs of the Theorem iterates through the combination process step by step (executing the fold).

The key point required to make the conditions of the theorem work for COFO processes is that the repeated combinatory processes are foldable.   This is achieved if they are mutually associative, as in that case series of repeated combination can be carried out step by step in any order, including the order implicit in a certain fold operation.    Given that the series of combinatory operations on the lhs decomposes into a fold (which is the case if the operations are mutually associative), then we can do fold fusion and we have made the "easy part" $\llparenthesis S \rrparenthesis \cdotp  \llparenthesis T \rrparenthesis^\circ$ of the specification of the dynamic programming process a fold.   We thus obtain, as a corollary of Theorem \ref{thm:DP}, the result that

\begin{thm}
A COFO DDS whose combinatory operations $C_i$ are mutually associative can be implemented as a chronomorphism.
\end{thm}

In the PLN inference context, for example, the approach to PLN inference using relaxation rather than chaining outlined in \cite{Goertzel2008prob} is one way of finding the fixed point of the recursion.    What the theorem suggests is that folding PLN  inferences across the knowledge metagraph is another way, basically boiling down to forward and backward chaining as outlined above -- but as we have observed above, this can only work reasonably cleanly for crisp inference if one uses PLN rules formulated as co-implications rather than one-way implications.

When dealing with the uncertainty-management aspects of PLN rules, one is no longer guaranteed associativity merely by adopting reversibility of individual inference steps, and one is left with a familiar sort of heuristic reasoning: One tries to arrange one's inferences as series of co-implications whose associated distributions have favorable independence relationships.   For instance if one is trying to fold forward through a series of probabilistically labeled co-implications, one will do well if each co-implication is independent of its ancestors conditional on its parents (as in a Bayes net); this allows one to place the parentheses in the same place the fold naturally does.   The ability of chronomorphisms to fulfill the specifications implicit in the relevant Galois connectiosn becomes merely an approximate heuristic guide, though we suspect still a very valuable one.

\subsection{Cognitive Manifestations of Implicit Programming}

An interesting variation on these themes is provided by the concept of "implicit programming" from Patterson's thesis on Constructible Duality logic \cite{patterson1998implicit}.   The basic idea is that sometimes when the least fixed point of an equation (defined in CD logic) doesn't exist, there is a notion of an "optimal fixed point" that can be used instead.  Roughly, if there is only one assignation of values to variables that could possibly serve as a fixed point, then this strategy assumes that it actually is the fixed point -- whether or not it can consistently be described as minimal.

The implication of this concept for our present considerations is that: Given a reversible paraconsistent proof, we can look at both the least and optimal fixed point of the Dynamic Programming equation (the rhs in Theorem \ref{thm:DP} above) corresponding to the proof as being equivalent to the proof.  The former gives the direct demonstration and the latter gives a derivation based on negated information.

The reason this is interesting in general is that sometimes least fixed point is undefined, but the optimal fixed point might still be defined.  For instance, sometimes when the consistency condition of the DP theorem fails, there could still be a solution via implicit programing.  This might occur when there is some circular reasoning involved for example -- a situation worthy of further analysis to be deferred to a later paper.  There may be a connection with reasoning about uncertain self-referential statements such as we have previously modeled using infinite-order probabilities \cite{InfiniteOrderProbabilities} -- statements like "This sentence is false with probability .7", reformulated in terms of CD logic, may  serve as toy examples where implicit programming yields intuitively sensible answers via optimal rather than minimal fixed points.

\section{Potential Connections to Quantum Computing and Cognition}

As a brief aside to our core considerations here, it is interesting to explore the potential extensions of the above ideas from the classical to the quantum computing domain.   The foundations of such extension may be found in the observation that the basic dynamic programming functional equation is mappable isomorphically into the Schrodinger equation \cite{lindgren2019quantum} \cite{contreras2017dynamic} \cite{ohsumi1994nonlinear}.

Given this mathematics, the question of how to frame the optimal-control interpretation of the Schrodinger equation in a cognitive science and AGI context becomes interesting.  One begins with the result that, under appropriate assumptions, minimizing action in quantum mechanics is isomorphic to minimizing expected reward in dynamic programming.   Conceptually one then concludes that

\begin{itemize}
\item At the classical-stochastic-dynamics level, one views the multiverse as being operated by some entity that is choosing histories (multiverse-evolution-paths) with probability proportional to their expected action (where expected action serves the role of expected reward).  
\item Then at the quantum level, one views the multiverse as being operated by some entity that is assigning histories amplitudes proportional to their expected action (where expected action serves the role of expected reward).  The amplitudes are summed and the stationary-action histories include those that in the "classical limit" give you the maximal-expected-action paths.
\end{itemize}

From a quantum computing view, if one has a classical algorithm that operates according to dynamic programming with a particular utility function $U$, one would map it into a quantum algorithm designed so that expected action maximization corresponds to expected $U$ maximization.   

For instance, suppose we start on the classical side with the time-symmetric Bellman's equation briefly discussed above.  Then on the quantum level we seem to obtain something similar to the representation used in the Consistent Histories interpretation \cite{griffiths1984consistent}.   That is, the forward and backward operators composed in the consistent-histories class operator appear to correspond to a solution of the fix-point equation corresponding to time-symmetric dynamic programming.   Many other connections arise here, e.g. in a quantum computing context

\begin{itemize}
\item where graphtropy is invoked above in defining reward functions for inference processes, one can potentially insert quangraphtropy \cite{Graphtropy} instead
\item  where standard PLN inference formulas manipulate real probabilities, one can substitute similarly defined formulas manipulating complex-valued amplitudes
\end{itemize}

\noindent  One sees here the seeds of a potential approach to deriving quantum cognition algorithms via Galois connections, but further validation of this potential will be left for later research.

\section{Implications for AGI-Oriented Programming Language Design}

In \cite{goertzel2020kind} it is asked what sort of programming language best suits implementation of algorithms and structures intended for use in an integrative AGI system.   Among the conclusions are that one probably wants a purely functional language with higher-order dependent types and gradual typing; and that the bulk of processing time is likely to be taken up with pattern-matching queries against a large dynamic knowledge-base, which means e.g. that computational expense regarding type-system mechanics is not an extremely sensitive point.

The two prequels to this paper \cite{goertzel2021paraconsistent} \cite{Goertzel2020metagraph} add some particulars to these conclusions, e.g. that an AGI language should include

\begin{itemize}
\item Efficient implementation of folding and unfolding across metagraphs, including
\begin{itemize}
\item Memory-carrying folds and unfolds, i.e. histomorphisms, futumorphisms, chronomorphisms
\item Continuation-passing-style implementations of all these folds and unfolds, to elegantly enable pause and restart within a fold
\end{itemize}
\item First and second order probabilistic types
\item Paraconsistent types, implemented via an individual node or edge having separate type expressions corresponding to positive vs. negative evidence
\end{itemize}

The explorations pursued here add a few more points to the list, e.g.:

\begin{itemize}
\item Simple tools for expressing DDSs, including those to be pursued via greedy, dynamic programming, or stochastic/approximate dynamic programming
\item Simple, efficient tools for sampling from distributions defined across typed metagraphs (as the COFO DDSs outlined here mainly rely on sampling of this nature in their decision step)
\item Simple tools for estimating measurements of probability distributions defined over typed metagraphs, e.g. entropy, logical entropy, graphtropy
\item Syntactic and semantic tools  for handling reversible operations, e.g. reversible logical inference rules and reversible program execution steps
\begin{itemize}
\item This entails some work specific to particular cognitive algorithms, e.g. figuring out the most pragmatically convenient way to represent common PLN inference rules as reversible rules
\end{itemize}
\item Tools making it easy to generate and maintain subpattern hierarchies, and to coordinate subpattern hierarchies and the hierarchies implicit in the optimization of DDSs using dynamic programming type approaches
\item Tools making it easy to express algebraic properties of sets of combinatory operations, e.g. mutual associativity
\item Ability to concisely express derivations of specific algorithms from case-specific instantiations of theorems such as the Greedy Theorem and Dynamic Programming Theorem from \cite{mu2012programming} presented and leveraged above.
\begin{itemize}
\item This provides a relatively straightforward path to enabling the system to derive new algorithms optimized for its own utilization in particular cases.
\end{itemize}
\end{itemize}

Putting these various pieces together, in the context of the OpenCog Hyperon (next-generation OpenCog) project, we are getting quite close to an abstract sort of requirements list for the deep guts of an Atomese 2 interpreter.

\section{Conclusion and Future Directions}

We have presented here a common formal framework for expressing a variety of AGI-oriented algorithms coming from diverse theoretical paradigms and practical traditions.   This framework is neither so specialized as to allow only highly restrictive versions of the algorithms considered, nor so general as to be vacuous (e.g. the AGI-oriented algorithms we find of interest can be represented especially concisely and simply in this framework, much more so than would be the case for arbitrary computational algorithms).  

Our explorations here have led to some interesting and nontrivial formal conclusions, e.g. regarding the power of the requirement of mutual associativity among combinatory operations in a combinatory-operation-based function-optimization setting.   Mutual associativity, it turns out, both leads to correspondence between subgoal hierarchies in goal-oriented processing and subpattern hierarchies in a knowledge story, and enables execution of  combinatory-operation-based function-optimization via futumorphisms and histomorphisms.   

On the other hand, it is also clear there is a heuristic aspect to the application of these theoretical results to practical AGI systems -- because in the case of real AGI systems, the knowledge base (e.g. the OpenCog Atomspace) is constantly changing as a result of cognitive actions, so that one usually can't precisely calculate a fold of a cognitive operator across a knowledge base without the knowledge base changing midway before the fold is done.   This is not necessarily problematic in practice, but it does mean that most of the theoretical conclusions drawn here are only precisely applicable in the limiting case where real-time knowledge base changes in the course of a fold/unfold operation (or other sort of cognitive operation) are minimal.  

Formal treatment of the general case of greedy or approximate stochastic DP style execution of a DDS in the case where many DDS action steps entail significant knowledge base revisions  remains a step for the future.   Even without that, however, it appears that the formulation given here does provide some clarity regarding the parallels and interrelationships between different cognitive algorithms, and the sort of infrastructure needed for effectively implementing these algorithms.

\bibliographystyle{alpha}
\bibliography{bbm}

\end{document}